\documentclass[conference, 10 pt]{ieeeconf}  

\IEEEoverridecommandlockouts                              

\usepackage{subcaption}
\usepackage{cite}

\usepackage{graphicx} 
\usepackage{tikz}
\usepackage{pgfplots}
\usepackage{amsmath} 
\usepackage{amssymb}  
\usepackage{import}
\usepackage{dsfont}
\usepackage{hyperref}
\usepackage{gensymb}
\usepackage{booktabs}
\usepackage{tabularx}
\usepackage{diagbox}

\title{\LARGE \bf
Dynamic Occupancy Grid Mapping\\with Recurrent Neural Networks
}

\author{Marcel Schreiber$^{1}$, Vasileios Belagiannis$^{1}$, Claudius Gl\"aser$^{2}$ and Klaus Dietmayer$^{1}$
	\thanks{The authors are with:}%
	\thanks{$^{1}$Institute of Measurement, Control, and Microtechnology, Ulm University, Germany, {\tt\footnotesize \{first.last\}@uni-ulm.de}}
	\thanks{$^{2}$Robert Bosch GmbH, Corporate Research, 71272 Renningen, Germany, {\tt\footnotesize \{first.last\}@de.bosch.com}}%
}
\begin{document}

\bstctlcite{IEEEexample:BSTcontrol} 

\def\cred{\textcolor{red}}
\def\cblue{\textcolor{blue}}
\def\cgreen{\textcolor{green}}

\newcommand\copyrighttextinitial{%

	\scriptsize This work has been submitted to the IEEE for possible publication. Copyright may be transferred without notice, after which this version may no longer be accessible.}%
\newcommand\copyrighttextfinal{%
	
	\scriptsize\copyright\ 2021 IEEE. Personal use of this material is permitted. Permission from IEEE must be obtained for all other uses, in any current or future media, including reprinting/republishing this material for advertising or promotional purposes, creating new collective works, for resale or redistribution to servers or lists, or reuse of any copyrighted component of this work in other works. DOI: 10.1109/ICRA48506.2021.9561375.}%
\newcommand\copyrightnotice{%

	\begin{tikzpicture}[remember picture,overlay]%

	\node[anchor=south,yshift=10pt] at (current page.south) {{\parbox{\dimexpr\textwidth-\fboxsep-\fboxrule\relax}{\copyrighttextfinal}}};%
	\end{tikzpicture}%


}

\newcommand{\ts}[1]{{\textsubscript{#1}}}
\newcommand{\tbf}[1]{{\textbf{#1}}}

\maketitle
\copyrightnotice%
\thispagestyle{empty}
\pagestyle{empty}

\begin{abstract}
Modeling and understanding the environment is an essential task for autonomous driving. 
In addition to the detection of objects, in complex traffic scenarios the motion of other road participants is of special interest. 
Therefore, we propose to use a recurrent neural network to predict a dynamic occupancy grid map, which divides the vehicle surrounding in cells, each containing the occupancy probability and a velocity estimate.
During training, our network is fed with sequences of measurement grid maps, which encode the lidar measurements of a single time step.
Due to the combination of convolutional and recurrent layers, our approach is capable to use spatial and temporal information for the robust detection of static and dynamic environment.
In order to apply our approach with measurements from a moving ego-vehicle, we propose a method for ego-motion compensation that is applicable in neural network architectures with recurrent layers working on different resolutions.
In our evaluations, we compare our approach with a state-of-the-art particle-based algorithm on a large publicly available dataset to demonstrate the improved accuracy of velocity estimates and the more robust separation of the environment in static and dynamic area.
Additionally, we show that our proposed method for ego-motion compensation leads to comparable results in scenarios with stationary and with moving ego-vehicle. 
\end{abstract}
%
%
\section{Introduction}
For a safe maneuver planning of self-driving vehicles, a robust perception and modeling of the environment is a crucial task.
In the recent years, significant process has been made in developing more accurate and robust object detection methods, mainly driven by deep learning approaches \cite{DiFengSurveyTransactions}.
However, all these methods rely on the object categories seen during training.
An alternative representation of the environment are occupancy grid maps, in which the vehicle surrounding is divided in equally sized grid cells, each containing the occupancy probability of the space its located \cite{Thrun:2005:PR:1121596}.
Despite the early usage of occupancy grid maps for mapping under the assumption of a static environment \cite{ElfesOccGrids}, in the setting of autonomous driving the vehicle surrounding is highly dynamic and moving objects are of special interest.
In this context, several works \cite{DanescuParticleBasedOccupancyGrid, Negre_HybridSamplingBayesian, TanzmeisterGridBasedMapping, DBLP:journals/corr/NussRTYKMGD16} propose to extend the cell state in occupancy grid maps with velocity estimates, represented by particles located in this cell.
The benefits of these dynamic occupancy grid maps (DOGMs) are the ability to represent arbitrary shaped objects and to estimate the motion of objects without the need of explicit data association.
However, the aforementioned particle-based approaches for estimating DOGMs come with mainly two drawbacks: the assumption of independent cells in the update step of the particle filter and the high computational effort due to the particle approximation.
\begin{figure}
	\centering
	\includegraphics[width=\columnwidth, trim=0 0 0 0, clip]{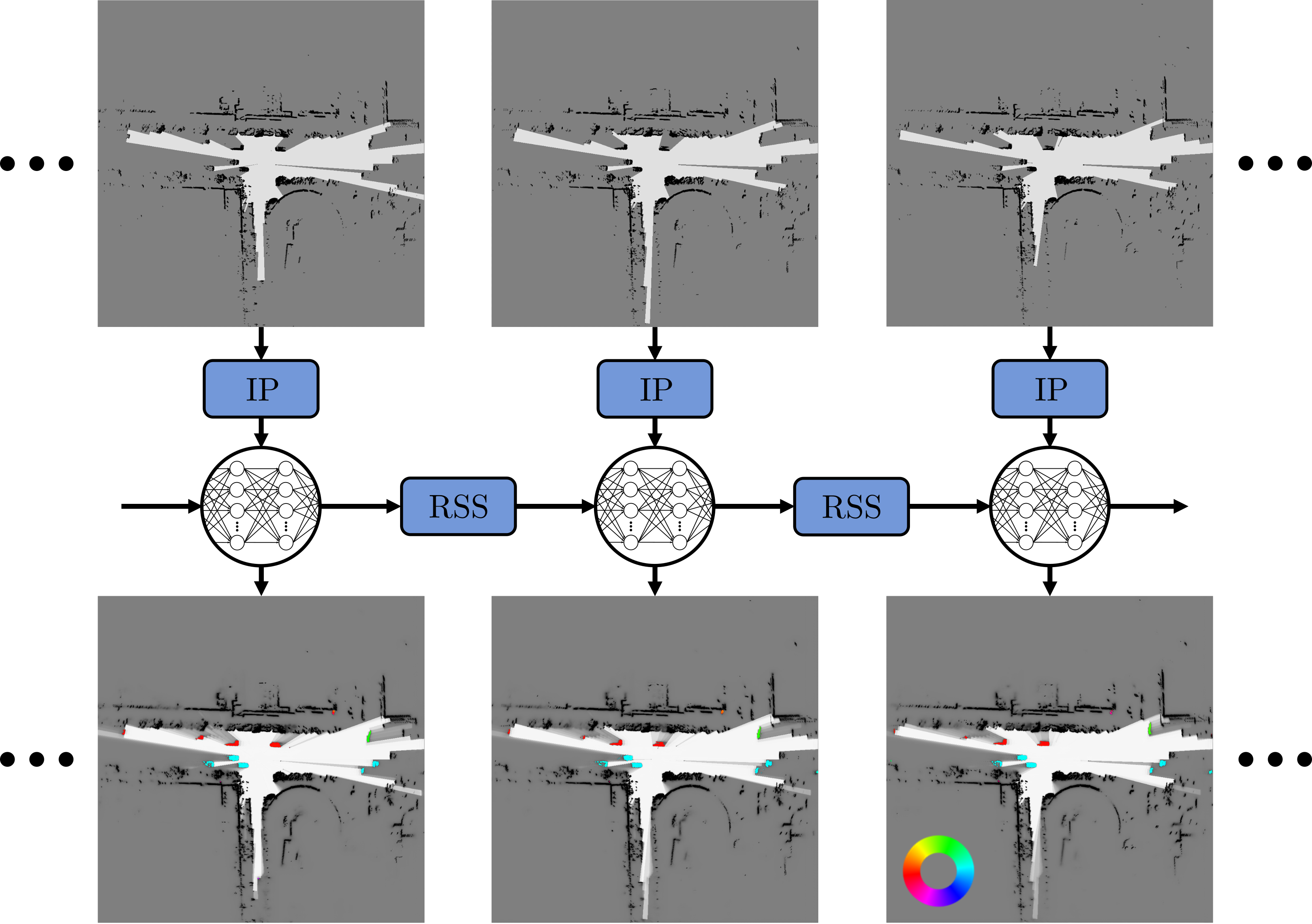}
	\caption{Illustration of our approach. The recurrent network predicts a dynamic occupancy grid map based on the current measurement grid map and recurrent states of the last time step. The ego-motion is compensated through a combination of input placement~(IP) and recurrent states shifting~(RSS). The occupancy probability is shown in gray scale with black for occupied cells and white for free space, the orientation of velocity estimates is encoded in colors.}
	\label{fig:SystemOverview}
\end{figure}
To tackle these drawbacks, we have proposed in prior work \cite{SchreiberMotionEstimationInOccupancyGrids} a learning-based approach for predicting DOGMs in a setting with stationary ego-vehicle.
Here, we have shown, that a learning-based approach provides more accurate velocity estimates in dynamic driving situations, e.g. a braking or turning car, and is able to reduce the erroneous velocity estimates in static environment by using spatial context. \\
In this work, we present based on \cite{SchreiberMotionEstimationInOccupancyGrids} a recurrent neural network (RNN) architecture for modeling the vehicle environment as DOGM in various scenarios with moving ego-vehicle.
The used network architecture is based on the U-Net \cite{DBLP:journals/corr/RonnebergerFB15}, but consists of several convolutional long short-term memories \cite{DBLP:journals/corr/ShiCWYWW15}, which are placed on different network levels to make use of temporal information on different resolutions.
As shown in Fig. \ref{fig:SystemOverview}, during deployment the recurrent neural network predicts one DOGM each time step, based on the current lidar measurements, encoded in a measurement grid map, and the recurrent states of the last time step.
In the depicted sequence with a moving ego-vehicle, the predicted DOGM show the ability to estimate static and dynamic area, decoupled from the ego-motion. 
In order to achieve this, we propose a combination of input placement (IP) to compensate small movements at the network input and recurrent states shifting (RSS), where all recurrent states in the network are shifted synchronously if a larger displacement over several time steps occurs.
Our evaluations show, that using this ego-motion compensation method, the system performs with the same accuracy for a stationary and a moving ego-vehicle.
Additionally, we compare our results with a particle-based algorithm in various driving situations, showing that we achieve more accurate velocity estimates, a more robust detection of static and dynamic cells and an improved usability for subsequent processing, e.g. clustering to detect moving objects. \\
To sum up, our contributions are \textit{1) the introduction of IP and RSS to apply our network with a moving ego-vehicle, 2) thorough evaluation on a large public dataset, 3) superior performance compared to a particle-based approach.}
\section{Related Work}
\noindent \textbf{Occupancy grid maps} are a popular environment representation in robotics for dividing the surrounding of an agent in an equally spaced grid, where each cell contains the occupancy state of the space its located \cite{ElfesOccGrids, Thrun:2005:PR:1121596}.
In the field of autonomous driving, several works \cite{DanescuParticleBasedOccupancyGrid, Negre_HybridSamplingBayesian, TanzmeisterGridBasedMapping, DBLP:journals/corr/NussRTYKMGD16} propose to extend the occupancy grid map with velocity estimates.
Danescu \textit{et~al.}~\cite{DanescuParticleBasedOccupancyGrid} introduce the particle-based occupancy grid maps, in which particles, consisting of position and velocity can freely move between cells and are used to describe their dynamic state.
The work in \cite{Negre_HybridSamplingBayesian} and \cite{TanzmeisterGridBasedMapping} adopted this approach and independently propose to only represent the dynamic cells with particles, whereas Nuss \textit{et~al.}~\cite{DBLP:journals/corr/NussRTYKMGD16} suggest to define the dynamic grid mapping as a random finite set problem.
Recently, several deep learning approaches have utilized grid maps as input data for convolutional neural networks to detect objects \cite{HoermannDetectionOnDynamicGridMaps, WirgesObjectDetectionAndClassification}, separate the occupied area in static or dynamic cells \cite{PiewakDynamicObjectDetection}, predict future occupancy \cite{HoermannDynamicOccupancyGridPrediction} or use self-supervised scene flow prediction to estimate odometry \cite{WirgesSelfSupervisedFlowEstimation}.
Dequaire \textit{et~al.}~\cite{DequaireDeepTrackingInTheWild} propose an end-to-end trainable recurrent neural network to estimate unoccluded occupancy grid maps, based on raw lidar data as input, but without having velocity estimates.
In contrast, Schreiber \textit{et~al.}~\cite{SchreiberMotionEstimationInOccupancyGrids} use a combination of convolutional and recurrent network layers to estimate both, filtered occupancy and velocity estimates for each cell, i.e.~a DOGM, similar to \cite{DBLP:journals/corr/NussRTYKMGD16}.  
This learning-based approach shows clear improvements compared to particle-based approaches, but only works for a stationary setting. In this work, we further develop this approach and propose our method for scenarios with moving ego-vehicle. 
In a concurrent approach, Filatov \textit{et~al.}~\cite{AnyMotionDetector} propose a novel architecture to estimate class-agnostic scene dynamics as grid representation using a sequence of lidar point clouds as input data, which is first processed in voxel feature encoding layers \cite{VoxelNet}.
Afterwards these features are aggregated in a convolutional recurrent network layer with ego-motion compensation and the last hidden state is processed in a ResNet18-FPN backbone network to finally predict a segmentation and velocity grid.
Their proposed ego-motion compensation is similar to \cite{DequaireDeepTrackingInTheWild}, but only suitable for network architectures, in which all recurrent layers are on the same level, i.e. using features with the same resolution. 
Recently, Wu \textit{et~al.}~\cite{MotionNet} propose a spatio-temporal pyramid network using a sequence of lidar data encoded as bird's eye view maps as input to classify cells and predict their future trajectory. 
In contrast to our approach, the ego-motion is compensated by transforming all past lidar point clouds to the current one, before feeding them into the network. This is possible, as they use a sequence of input data for every new prediction during deployment. 
\\
\noindent \textbf{Scene Flow Estimation} describes the task to estimate the motion in a scene represented as three-dimensional flow vectors, which describe the correspondence between points of consecutive time steps~\cite{ThreeDimensionalSceneFlow, ShapeAndNonerigidMotionEstimationThrough}.
Dewan \textit{et~al.}~\cite{RigidSceneFlowFor3DLidarScans} propose to estimate scene flow in 3D lidar scans by formulating it as an energy optimization problem, relying on SHOT \cite{tombari2010unique} feature descriptors.
In contrast, Ushani \textit{et~al.}~\cite{ALearningApproachForRealTimeTemporalSceneFlow} propose to first generate occupancy grids from lidar scans and then apply a learned background filter. 
Then an expectation-minimization algorithm, which leverages an occupancy constancy metric is used to compute scene flow.
A drawback of these classical approaches is the reliance on a handcrafted feature or metric.
The work in \cite{pmlr-v87-ushani18a} is built on \cite{ALearningApproachForRealTimeTemporalSceneFlow} and uses a learned feature space instead of the occupancy constancy metric, but loses the real-time capability.
Recently, several works \cite{FlowNet3D, DeepParametricContinuousConvolutionalNeuralNetworks, HPLFlowNet, PointFlowNet} have proposed deep neural network approaches to estimate 3D scene flow in lidar point clouds.
Liu \textit{et~al.}~\cite{FlowNet3D} introduce the \textit{FlowNet3D}, which is based on the \textit{PointNet++} architecture \cite{qi2017pointnet++}, and extend it with the novel flow embedding layer to associate points between a pair of consecutive point clouds.
This approach is further refined in \cite{FlowNet3D++} by incorporating geometrically constraints to the network optimization.
In contrast, the \textit{HPLFlowNet} \cite{HPLFlowNet} use an alternative architecture inspired by \textit{Bilateral Convolutional Layers} \cite{Jampani2016LearningSH} to more efficiently process pairs of point clouds.
The \textit{PointFlowNet} \cite{PointFlowNet} groups two consecutive lidar point clouds in voxels and generates feature maps as proposed in \cite{VoxelNet}. 
These feature maps are used in several decoder branches to jointly estimate scene flow, predict rigid motion, and detect objects.
Wang \textit{et~al.}~\cite{DeepParametricContinuousConvolutionalNeuralNetworks} introduce a parametric continuous convolution operation, which is a learnable kernel function that can handle unstructured data such as 3D point clouds. 
All the aforementioned methods rely on raw 3D lidar data and focus on the prediction of accurate correspondences between two consecutive point clouds.
Moreover, most of these approaches \cite{FlowNet3D, FlowNet3D++, PointFlowNet} are not real-time applicable.
In contrast, we choose to work with a grid map representation and estimate motion as 2D vectors, assuming only dynamics on the ground plane, which seems reasonable in an autonomous driving setting. 
A more similar data representation is used in \textit{PillarFlow} \cite{lee2020pillarflow}, where an end-to-end method is introduced to estimate 2D motion in a lidar bird's eye view image.
Here, two consecutive lidar point clouds are encoded as proposed in \cite{lang2019pointpillars}, these 2D representations are then used as input for a flow network based on \cite{PWCNet} to estimate 2D flow.
However, in our approach we predict grid maps with a size of $1001\times1001$ cells, which is equal to $150m\times150m$, providing a significantly larger field of view, than the representation used in \cite{lee2020pillarflow}. 
In contrast to all mentioned scene flow approaches, which rely on two consecutive point clouds, we propose a RNN architecture using information from more than only two time steps.
\section{Methods}\label{sec:methods}
In this section, we present our learning-based approach to estimate DOGMs.
First, we formulate the problem of estimating DOGMs and summarize our prior work \cite{SchreiberMotionEstimationInOccupancyGrids}, which is the basis for the network architecture of our approach.
Next, we introduce our approach for ego-motion compensation to apply our network architecture in scenarios with a moving ego-vehicle.
\subsection{Problem Formulation}
We aim to estimate DOGMs based on a sequence of measurement grid maps.
A grid map divides the vehicle surrounding in individual cells with equal edge size $a$, where each cell contains information of the space where its located.
In occupancy grid maps, each cell is described by a state $o_k$, to be free or occupied \cite{ElfesOccGrids, Thrun:2005:PR:1121596}.
Measurement grid maps are occupancy grid maps, based on sensor measurements $z_k$ of a single time step $k$.
So, the vehicle environment is represented as a tensor in $\mathbb{R}^{W\times H\times c\textsubscript{m}}$ with the width $W$, the height $H$ and one channel $c\textsubscript{m}=\{p_{z_{k}}\}$ containing the single frame occupancy probability $p_{z_k}$ for each cell.
These measurement grid maps serve as input data of our model and are calculated as described in \cite{SchreiberMotionEstimationInOccupancyGrids}.
The output of our approach is a DOGM with the same spatial size and the channels $c\ts{d} =\{p\ts{o}, v\ts{E}, v\ts{N}\}$, containing the filtered occupancy probability $p\ts{o}$ and velocities pointing east $v\ts{E}$ and north $v\ts{N}$.
\subsection{Network Architecture in Stationary Setting}\label{sec:NetworkSetup}
\begin{figure}
	\centering
	\vspace{1.42mm}
	\includegraphics[width=\columnwidth]{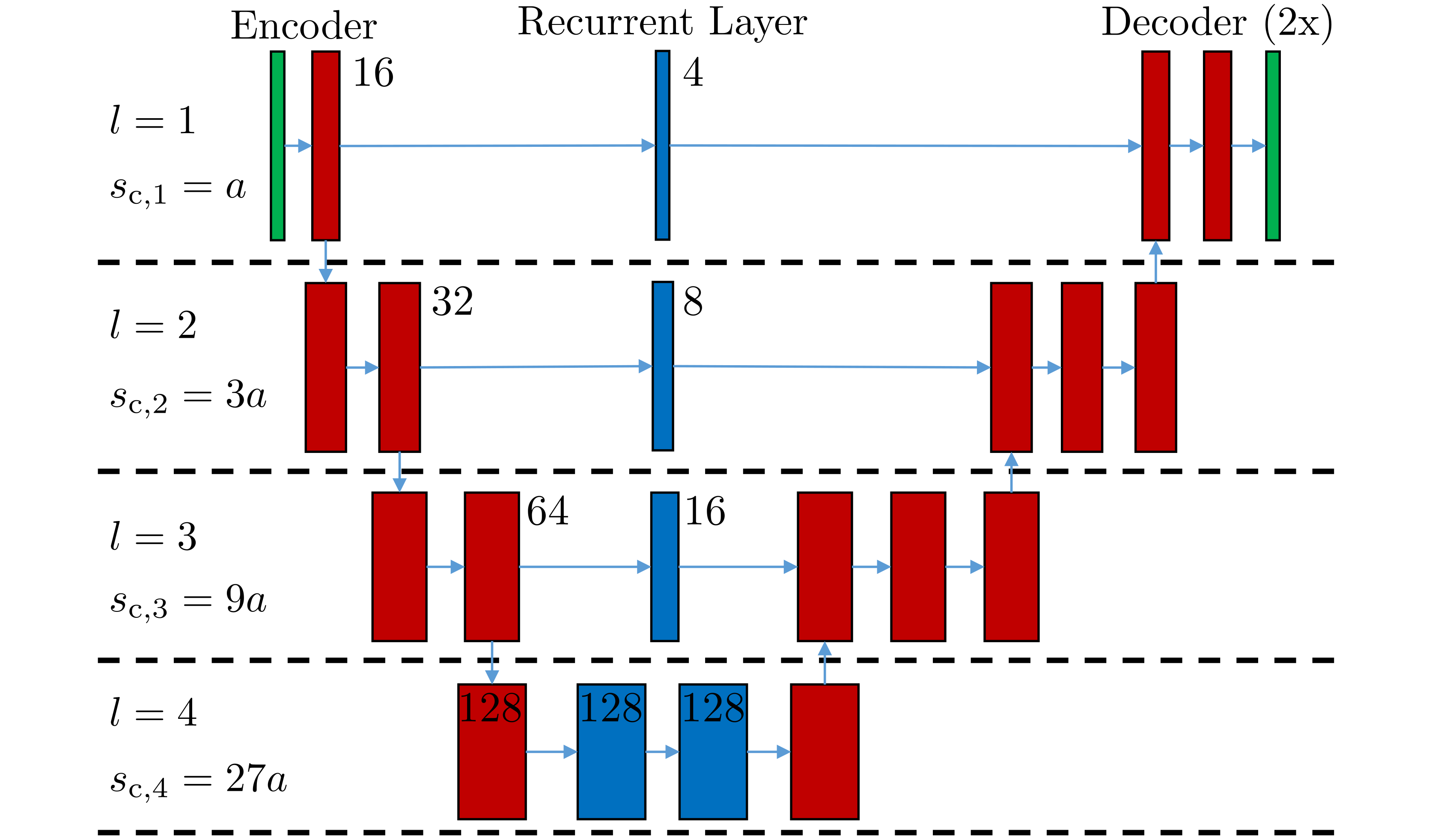}
	\caption{Illustration of our network architecture with feedforward layer in red and recurrent layer in blue. 
	}
	\label{fig:network}
	\vspace{-1.42mm}
\end{figure}
Our network architecture, shown in Fig.~\ref{fig:network} combines feedforward and recurrent layers similar to \cite{SchreiberMotionEstimationInOccupancyGrids}. 
During training a sequence of $n\ts{in}$ measurement grid maps, with dimensions $\mathbb{R}^{n\ts{in}\times1001\times 1001\times 1}$, are processed by the encoder, which consists of several convolutional layers and reduces the input three times using convolutions with a stride of 3 to a representation of $\mathbb{R}^{n\ts{in}\times38\times 38\times 128}$.
This tensor is used as input for a two-layer convolutional long short-term memory (ConvLSTM) \cite{DBLP:journals/corr/ShiCWYWW15} with hidden and cell states of the same size as the input.
Additionally, we inserted ConvLSTMs in each skip connection as first proposed in \cite{SchreiberLongTermOccupancyGridPrediction}. 
In these three Conv\-LSTMs, we use hidden and cell states with less channels than the input data to save memory.
The described architecture leads to four network levels $l\in\{1,2,3,4\}$ with the grid cell sizes $\boldsymbol{s_{c}}=[s_{c,1}, s_{c,2}, s_{c,3}, s_{c,4}]^T = a\cdot[1, 3, 9, 27]^T$ with $a=0.15$~m.
The outputs of the recurrent layers are used in two separate decoders, which are built as mirrored encoders with transposed convolutional layers to scale the data to the same spatial size as the input. 
The first decoder is used to predict the occupancy probability $p\ts{o}$, the second decoder for the velocities $v\ts{E}, v\ts{N}$.
Compared to the architecture in \cite{SchreiberMotionEstimationInOccupancyGrids}, we added the classification of dynamic cells as an auxiliary task in the second decoder, as it improves the regression of the velocities $v\ts{E}, v\ts{N}$.
\subsection{Dynamic Grid Mapping with Moving Ego-Vehicle}\label{sec:EgoComp}
\begin{figure*}
	\centering
	\vspace{1.42mm}
	\includegraphics[width=\textwidth]{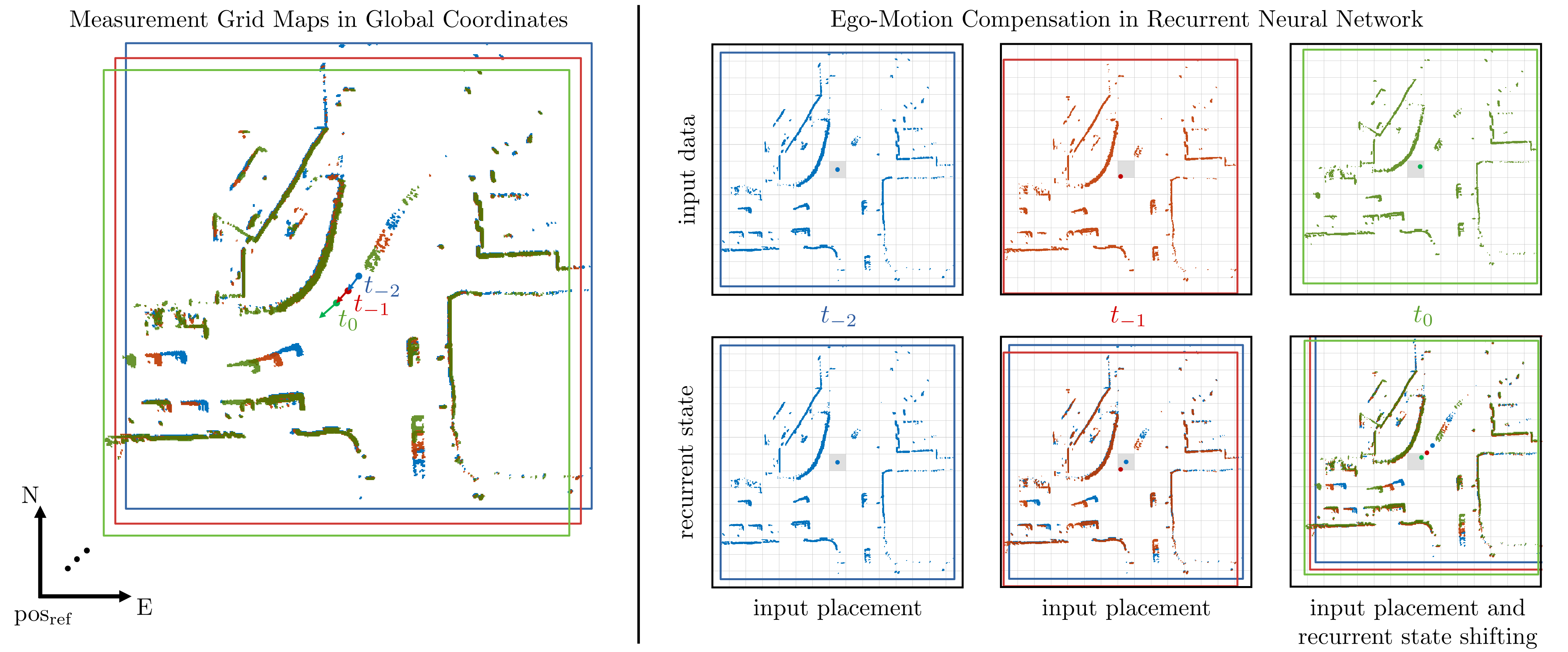}
	\caption{Illustration of our approach for ego-motion compensation. The images show the occupied cells in measurement grid maps, using colors to visualize the three time steps. Left: Visualization of input data in global coordinate system with the ego-pose marked as arrow. Top right: Input data placed relative to global grid. Bottom right: Recurrent states for the three time steps. On the right, a gray grid visualizes the grid cell sizes in the most inner recurrent layer. The ego-vehicle is always placed in the middle cell, marked in gray color.}
	\label{fig:ego_motion_comp}
	\vspace{-1.42mm}
\end{figure*}
In a setting with a non-moving ego-vehicle, stationary cells remain at the same location and the recurrent neural network, as described in Sec.~\ref{sec:NetworkSetup}, can directly detect moving areas and predict the global velocities $v\ts{E}, v\ts{N}$ for each cell.
To make the same predictions with a moving ego-vehicle, the ego-motion needs to be compensated by turning the moving setting back to the stationary setting inside our network architecture.
To make our motivation clear, we illustrate our approach for ego-motion compensation in Fig.~\ref{fig:ego_motion_comp} based on measurement grid maps of three consecutive time steps.
On the left side, the measurement grid maps are visualized in the global coordinate system. 
Note that the maps can be always placed in a global grid that is aligned in north and east direction where the grid origin is $\text{pos}\ts{ref}$ and the grid cell size is 0.15~m regardless of the orientation of the ego-vehicle.
In this global setting, the stationary areas lie on top of each other and the movement of objects is decoupled from the ego-motion.
A possibility to achieve this ego-motion compensated setting inside a recurrent layer is to shift the internal states according the movement of the ego-vehicle before updating them with the new input data.
However, in our network, we apply recurrent layers on different resolutions, so simply shifting the recurrent states would lead to inconsistencies between the recurrent states on different network levels.
To circumvent this problem, we propose to compensate movements inside a cell on level 4 directly at the input with our input placement (IP).
If a larger drift over several time steps occurs, i.e. a movement between cells on level 4, then all internal recurrent states are shifted synchronously with our recurrent states shifting (RSS).
For the ego-motion compensation, we use the current ego-vehicle position as an index $\boldsymbol{i_{\text{g},l}}$ in the global grid with the grid cell size $s_{\text{c},l}$, and the index in east $i_{\text{E},l}$ and north direction $i_{\text{N},l}$.
This global grid index is defined as 
\begin{equation}\label{eq:globalIndex}
	\boldsymbol{i_{g,l}} = [i_{\text{E},l}, i_{\text{N},l}]^T = \left\lfloor\left(\frac{\boldsymbol{\text{pos}\textsubscript{ego}}-\boldsymbol{\text{pos}\textsubscript{ref}}}{s_{c,l}}\right)\right\rfloor,
\end{equation}
with the position of the ego-vehicle in a global coordinate system $\boldsymbol{\text{pos}\textsubscript{ego}} = [E\textsubscript{ego}, N\textsubscript{ego}]^T$ for each network level $l$.
In Fig.~\ref{fig:ego_motion_comp}, we visualize the global grid with the grid cell size $s_{c,4}$ of the most inner recurrent layer, and marked the middle cell in gray, which has the global grid index $\boldsymbol{i_{g,4}}$.
As the global grids for each grid cell size $s_{c,l}$ have the same origin $\text{pos}\ts{ref}$, the relative position of the ego-vehicle inside this middle cell is
\begin{equation}\label{eq:inputPlacement}
	\boldsymbol{p\textsubscript{in}} = \boldsymbol{i_{g,1}} - \boldsymbol{i_{g,4}} \cdot \frac{s_{c,4}}{s_{c,1}}.
\end{equation}
In our IP the input data is increased with 28 cells per edge and $\boldsymbol{p\textsubscript{in}}$ is used in an adjustable padding operation inside the network to compensate small ego-motions at the input. 
As depicted in the example in Fig.~\ref{fig:ego_motion_comp}, the movement of the ego-vehicle between the time steps $t_{-2}$ and $t_{-1}$ is solely compensated with IP.
The movement between $t_{-1}$ and $t_0$ is compensated with a combination of IP and RSS, as the ego-vehicle changes its ego-cell $\boldsymbol{i_{g,4}}$.
The resulting grid difference $\boldsymbol{gd_{4,t_0}}$ is calculated with
\begin{equation}\label{eq:lowestShift}
	\boldsymbol{gd_{4,t_0}} = \boldsymbol{i_{g,4,t_0}} - \boldsymbol{i_{g,4,t_{-1}}}, 
\end{equation}
using the global grid index of the current time step $\boldsymbol{i_{g,4,t_0}}$, respectively of the previous time step $\boldsymbol{i_{g,4,t_{-1}}}$.
Accordingly, the corresponding grid differences $\boldsymbol{gd_{i,t_0}}$ for the other three network levels can be calculated as
\begin{equation}\label{eq:synchronShifts}	
	\boldsymbol{gd_{i,t_0}} = \boldsymbol{gd_{4,t_0}} \cdot \frac{s_{c,4}}{s_{c,i}}, \quad i = 1,2,3
\end{equation}
based on the largest grid cell size in the network $s_{c,4}$.
For each recurrent state of the last time step $\boldsymbol{h_{l,t_{-1}}},\boldsymbol{c_{l,t_{-1}}}$ we apply the transformation operation $f$, which performs a tensor shift based on the grid difference $\boldsymbol{gd_{l,t_0}}$ to get the new ego-motion compensated recurrent states 
\begin{equation}\label{eq:shiftStates}	
	\begin{aligned}
	\boldsymbol{\hat{h}_{l,t_{-1}}} &= f(\boldsymbol{h_{l,t_{-1}}}, \boldsymbol{gd_{l,t_0}}) \\
	\boldsymbol{\hat{c}_{l,t_{-1}}} &= f(\boldsymbol{c_{l,t_{-1}}}, \boldsymbol{gd_{l,t_0}}). 
	\end{aligned}
\end{equation}
This synchronous recurrent states shifting ensures, that the recurrent states in different network levels remain consistent, with the effect that our system behaves the same way for a moving and a stationary ego-vehicle, as visualized in Fig.~\ref{fig:ego_motion_comp}. \\
It is important to note, that both compensation methods, i.e. input placement and recurrent states shifting, are not sufficient to achieve an ego-motion compensation on its own for our architecture.
The solely shifting of the recurrent states, as applied in \cite{AnyMotionDetector, DequaireDeepTrackingInTheWild} is only applicable, if all recurrent layers have the same grid cell size.
We argue, that this is a strong limitation for the insertion of recurrent layers in fully convolutional network architectures, as most of them use network layers with different scales, e.g.~\cite{DBLP:journals/corr/RonnebergerFB15}, \cite{FeaturePyramidNetworks}. 
Moreover, the compensation of the ego-motion solely in the input data is only applicable, if a sequence of input data is used for each new prediction during deployment, as applied in \cite{MotionNet}, \cite{lee2020pillarflow}. 
Using a sequence of input data for every new prediction would lead to a high memory consumption, a slow inference time and counteracts the benefits of using recurrent neural networks for processing continuous data streams.
Unlike, we only use the current measurement grid map and the recurrent states of the previous time step to predict the current DOGM in our approach, which is possible due to the proposed ego-motion compensation method.
\subsection{Loss Function}
The loss function is a weighted sum of the occupancy loss $L\textsubscript{P\textsubscript{O}}$, the velocity loss in east $L\textsubscript{v,E}$ and north $L\textsubscript{v,N}$,  and the classification loss of dynamic cells $L\textsubscript{P\textsubscript{d}}$, defined as
\begin{equation}\label{eq:overallLoss}
L\textsubscript{o} = \alpha\textsubscript{p}L\textsubscript{P\textsubscript{O}} + \alpha\textsubscript{v}(L\textsubscript{v,E}+L\textsubscript{v,N}) + \alpha\textsubscript{d}L\textsubscript{P\textsubscript{d}},
\end{equation}
where the weight factors $\alpha\textsubscript{p}$, $\alpha\textsubscript{v}$ and $\alpha\textsubscript{d}$ determine the balance between the loss terms.
Here, we use the Huber-Loss with $\delta=0.02$ for the occupancy loss $L\textsubscript{P\textsubscript{O}}$ and a L2-Loss for the velocity losses $L\textsubscript{v,E}$, $L\textsubscript{v,N}$, as described in \cite{SchreiberMotionEstimationInOccupancyGrids}, with the same cell-wise weighting.
To include more supervision, we propose an auxiliary output to classify dynamic cells by using an L2-Loss for $L\textsubscript{P\textsubscript{d}}$ with the same cell-wise weighting as for the velocities.
The velocity regression and the classification of dynamic cells is only optimized in area of occupied cells where the threshold to define cells as occupied is $p\ts{o} > 0.7$.
In order to balance the sum of the loss terms we choose the factors $\alpha\textsubscript{p} = 50$, $\alpha\textsubscript{v} = 0.02$ and $\alpha\textsubscript{d} = 0.1$.
\section{Experimental Setup}
In our experiments, we use the publicly available \textit{Argoverse 3D Tracking Dataset} \cite{Argoverse} to investigate the performance of our approach in various complex driving situations.
Here, we discuss the generation of ground truth data on this dataset, and provide the implementation details of our model.
\subsection{Dataset}\label{sec:dataset}
For the training and evaluation of our approach we rely on the \textit{Argoverse 3D Tracking Dataset} \cite{Argoverse}, which consists of 113 recordings, 15 to 30 seconds each with accurate 3D track labels. 
The dataset is divided into 65 training, 24 validation and 24 test sequences, from which we use only the 65 training sequences for network optimization and the 24 validation sequences for evaluation as there are no labels for the test sequences.
The sensor setup includes among others two roof-mounted lidar sensors with 32 beams each, rotating with 10~Hz and a range up to 200~m, resulting in a point cloud of approximately 107,000 points per sweep.
We use the lidar data to calculate measurement grid maps as described in \cite{SchreiberMotionEstimationInOccupancyGrids}, which serve as the input to our network and the dynamic grid map algorithm in \cite{DBLP:journals/corr/NussRTYKMGD16}.
Here, we choose a grid cell size of 0.15~m and a grid map with $1001\times1001$ cells, leading to a perception area of [-75, 75]~m in east and north.
For our occupancy label, we use the occupancy probability $p\ts{o}$ of the particle-based DOGM in \cite{DBLP:journals/corr/NussRTYKMGD16}.
The ground truth velocities are obtained by calculating the velocities for each track label, using the displacement between frames in a global coordinate system.
We then associate the velocities to the occupied cells belonging to the object, according the bounding boxes.
\subsection{Implementation}
For the optimization, we use the ADAM optimizer \cite{DBLP:journals/corr/KingmaB14} with initial learning rate of $l\ts{r}=1e^{-4}$, and reduce it by a factor of 0.5 every 100k iterations.
During training, we calculate our loss based on the last two time steps and use truncated backpropagation through time \cite{AnEfficientGradientBasedAlgorithmRNN} with sequences of $n\ts{in} = 12$ measurement grid maps.
To improve our training process we employ data augmentation, i.e.~we rotate our grid maps randomly in $1\degree$ steps.
Additionally, we apply dropout in the non-recurrent connections in our ConvLSTM layers, as proposed in \cite{DBLP:journals/corr/ZarembaSV14}.
Note that we train our model based on data with a size of $601\times601$ due to memory constraints. During inference, though, we always make use of the full grid map size, that is  $1001\times1001$. This is possible because of our fully convolutional network architecture.
Our model achieves an inference time of about 53~ms on a Nvidia GeForce GTX 1080 Ti and therefore is real-time capable.
\section{Evaluation}
In our evaluations, we consider the validation set of the \textit{Argoverse Tracking Dataset}, consisting of 24 sequences, for which we calculate our ground truth data as described in Section \ref{sec:dataset}.
For comparison we use the state-of-the-art particle-based approach in \cite{DBLP:journals/corr/NussRTYKMGD16}, denoted as \textit{Nuss et~al.}, our RNN-based approach is stated as \textit{ours}.
First, we evaluate the ability to separate the occupied cells in static and dynamic, and the accuracy of the velocity estimates on cell level similar to scene flow approaches.
In addition, we apply a cluster algorithm to detect moving objects and evaluate both, the detection performance and the velocity estimates on object level. 
In the last part of our evaluation, we compare the performance in scenarios with a stationary and with a moving ego-vehicle quantitatively and therefore show the effectiveness of our approach for ego-motion compensation.
\subsection{Separation in Static and Dynamic Environment}
First, we evaluate the ability to divide the occupied cells in static and dynamic area.
Here, we mark cells with a velocity magnitude $v\ts{m}>0.8$~m/s as dynamic, all remaining occupied cells are static. 
In the particle-based approach each cell holds the mean velocity and the variance of the particle velocities, associated with this cell. 
Based on these values a mahalanobis distance $m$ between the mean velocity and zero can be calculated as described in \cite{DBLP:journals/corr/NussRTYKMGD16}.
For the particle-based approach, we use the mahalanobis distance to determine cells with $m^2<2.5$ as static cells, additionally to the velocity threshold above.
As metric, we calculate the mean intersection over union (mIoU), considering the two classes static and dynamic.
As shown in Table \ref{tab:CellLevel} our approach is superior in separating the occupied area in static and dynamic cells. 
Here, the neural network has the advantage to use context information in convolutional layer to reduce false positives in static area.
\begin{table}
	\centering
	\vspace{1.42mm}
	\setlength{\tabcolsep}{0.1cm}
	\caption{Evaluation on cell level}
	\label{tab:CellLevel}
	\begin{tabular}{@{}c|c|c|cccc@{}}
		\toprule
		sequences &method & mIoU& EPE\textsubscript{occ} &EPE\textsubscript{dyn} &EPE\textsubscript{slow}&EPE\textsubscript{fast}  \\ 
		& & & [m] & [m] & [m] & [m] \\
		\midrule
		all& Nuss \textit{et~al.} & 0.6247  & 0.0379  & 0.1998  & 0.1397 & 0.2197 \\
		&ours & \tbf{0.8892} & \tbf{0.0076} & \tbf{0.1059} & \tbf{0.0940} & \tbf{0.1099} \\		
		\midrule
		moving&Nuss \textit{et~al.} & 0.6032 & 0.0410 & 0.1975 & 0.1411 & 0.2168 \\
		&ours & \textbf{0.8826} & \textbf{0.0072} & \textbf{0.1041} & \textbf{0.0951} & \textbf{0.1073} \\
		\midrule
		stationary&Nuss \textit{et~al.} & 0.7793 & 0.0206 & 0.2083 & 0.1341 & 0.2302 \\
		&ours & \textbf{0.9155} & \textbf{0.0101} & \textbf{0.1126} & \textbf{0.0895} & \textbf{0.1197} \\
		\bottomrule
	\end{tabular}
	\vspace{-1.42mm}
\end{table}
\subsection{Evaluate Velocity on Cell Level}
For the evaluation of the velocity estimates on cell level, we use the end-point-error (EPE), which is the average L2 distance between estimated flow and ground truth, and a commonly used metric to measure the accuracy of scene flow.
We provide the EPE in different velocity ranges based on the velocity magnitude of the label $v\ts{m}$, considering all occupied cells with $p\ts{o}>0.7$ denoted as \textit{occ} and all occupied dynamic cells with $v\ts{m}>0.8$~m/s denoted as \textit{dyn}.
Additionally, we separate the dynamic cells in the ranges \textit{slow} with $0.8<v\ts{m}\leq3$~m/s and \textit{fast} $v\ts{m}>3$~m/s.
Note, that the EPE is calculated in meters and the time difference between two frames is 0.1 seconds.
The results in Table \ref{tab:CellLevel} show, that our model provides more accurate velocity estimates in all ranges.
\subsection{Evaluation on Object Level}
\begin{table}
	\centering
	\vspace{1.42mm}
	\setlength{\tabcolsep}{0.1cm}
	\caption{Evaluation on object level}
	\label{tab:ObjectLevel}
	\begin{tabular}{@{}c|cccccc@{}}
		\toprule
		method& MAE\textsubscript{vel}&MAE\textsubscript{ori}&$\bar{\sigma}\textsubscript{vel}$&$\bar{\sigma}\textsubscript{ori}$&Recall&Precision \\
			& [m/s] & [$\degree$] & [m/s] & [$\degree$] & & \\
		\midrule
		Nuss \textit{et~al.} & 0.5264 & \tbf{5.5752} & 0.4538 & 8.6861 & 0.8088 & 0.2734 \\
		ours & \tbf{0.5173} & 5.7394 & \tbf{0.4163} & \tbf{4.0315} & \tbf{0.8787} & \tbf{0.8108} \\
		\bottomrule
	\end{tabular}
	\vspace{-1.42mm}
\end{table}
Besides the evaluation on cell level, we evaluate our approach more application orientated.
Therefore, we apply a DBSCAN \cite{DBSCAN} algorithm using the position and velocities of each occupied dynamic cell and calculate the mean velocity and mean orientation of each cluster.
We consider each cluster with a mean velocity $v\ts{m}>0.8$~m/s as possible dynamic object, associate these clusters with the overlapping ground truth object box in a bird's eye view and calculate the recall and precision.
Then, we evaluate the accuracy of the velocity estimates for all true positives, by calculating the absolute error of the velocity magnitude and orientation for each object.
Additionally, we compute the standard deviations of the velocity magnitude and the orientation for each object as measurement of the extent in which the estimates of each cell belonging to one object differ.
Table \ref{tab:ObjectLevel} shows the mean errors of the velocity magnitude MAE\textsubscript{vel} and orientation MAE\textsubscript{ori} and the mean standard deviations for velocity magnitude $\bar{\sigma}\textsubscript{vel}$ and orientation $\bar{\sigma}\textsubscript{ori}$, considering all true positives.
The errors of the particle-based and the RNN-based approach are comparable, whereas the standard deviations are higher for the particle-based approach as visualized in Fig.~\ref{fig:qualitativeEvalStationary}.
This scene also illustrates that the RNN is clearly superior in estimating the velocity for turning vehicles, whereas the particle-based approach provides similarly accurate estimates for driving straight objects, which are predominant in most driving situations.
However, the recall as well as the precision are much higher with our proposed method, which makes it easier usable for subsequent processing, e.g. a fusion of the DOGM in an environment perception framework, as proposed in \cite{EnvironmentPerceptionFrameworkFusingMotDOGMaDM}.
The improved performance is mainly caused by the more accurate separation of slow moving area from static area, and the decreased amount of incorrect velocity estimates in static area.
\subsection{Ego-Motion Compensation}
\begin{figure}
	\vspace{1.42mm}
	\begin{subfigure}{\columnwidth}
		\includegraphics[width=0.49\columnwidth, trim=0 10 0 4, clip]{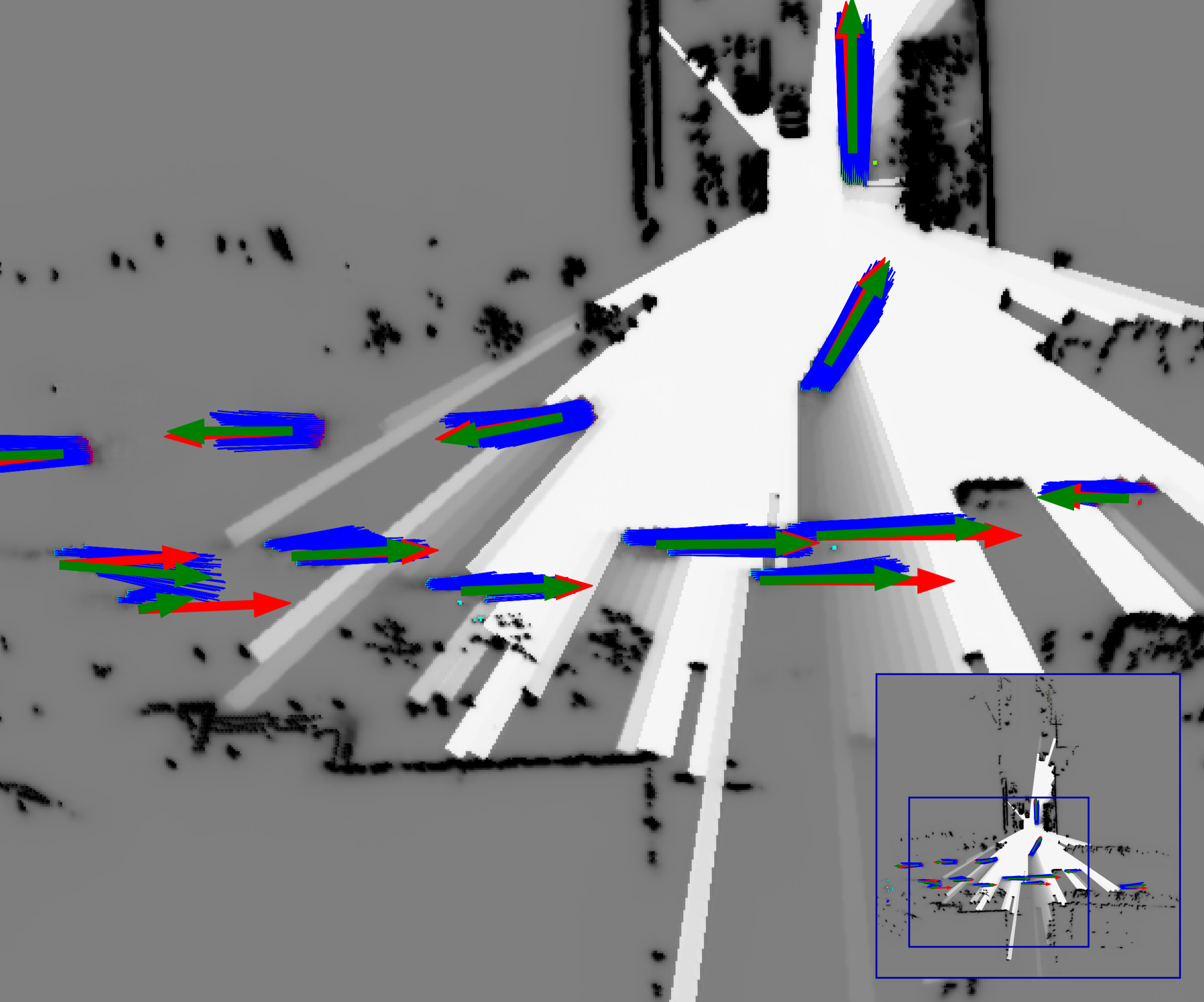}
		\includegraphics[width=0.49\columnwidth, trim=0 10 0 4, clip]{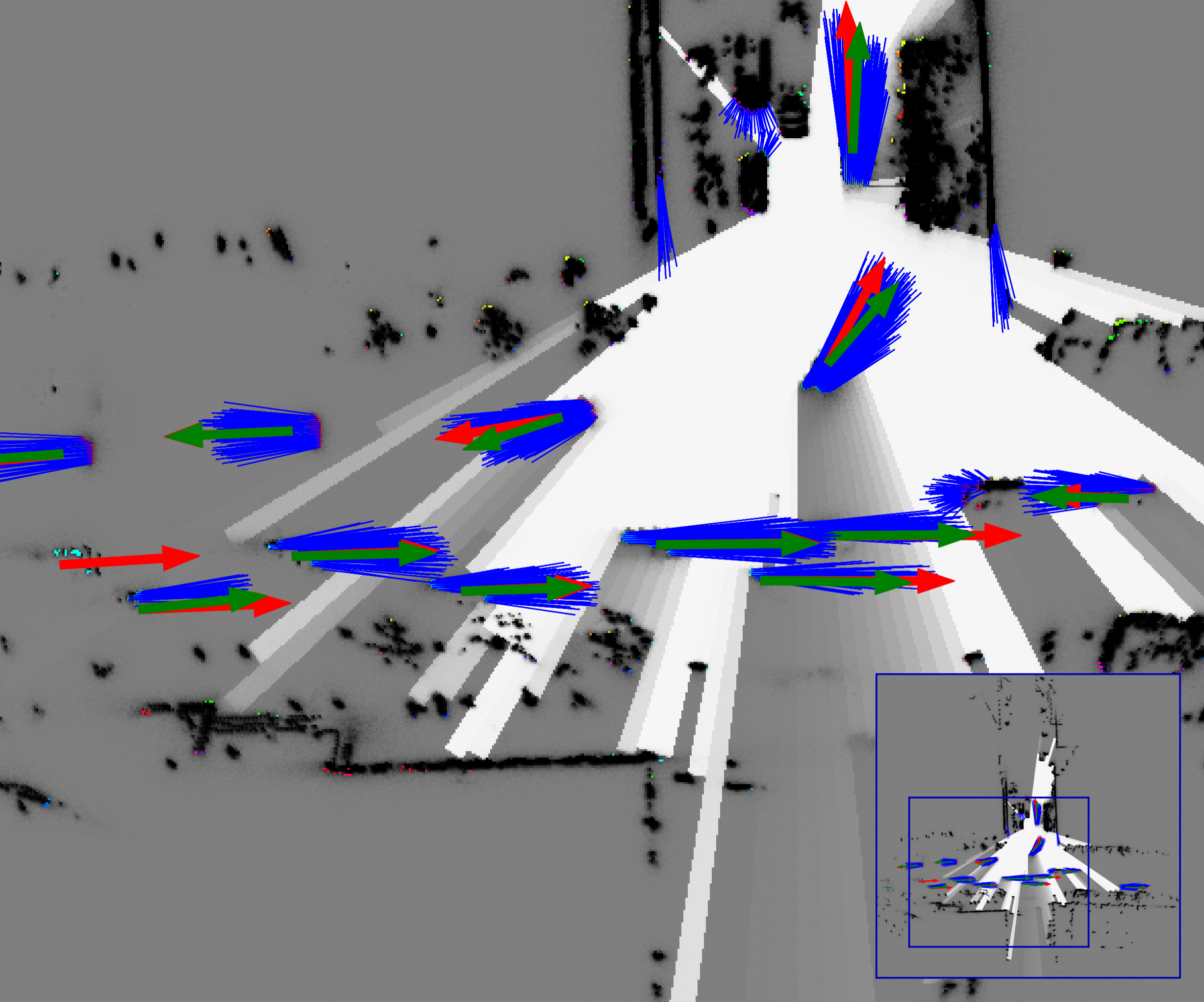}
		\subcaption{The velocity estimates are visualized as blue arrows for all detected dynamic object clusters. The mean velocities of clusters are depicted as green arrows, the ground truth as red arrows.}
		\label{fig:qualitativeEvalStationary}
	\end{subfigure}
	\begin{subfigure}{\columnwidth}
		\includegraphics[width=0.49\columnwidth, trim=20px 20px+2 0 6, clip]{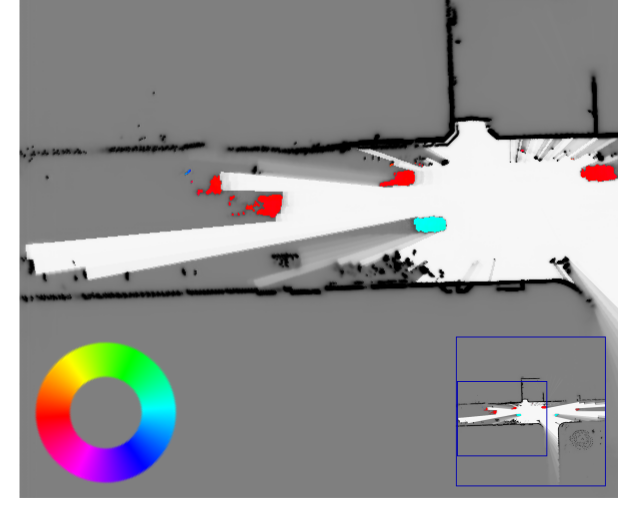}
		\includegraphics[width=0.49\columnwidth, trim=20px 20px+2 0 6, clip]{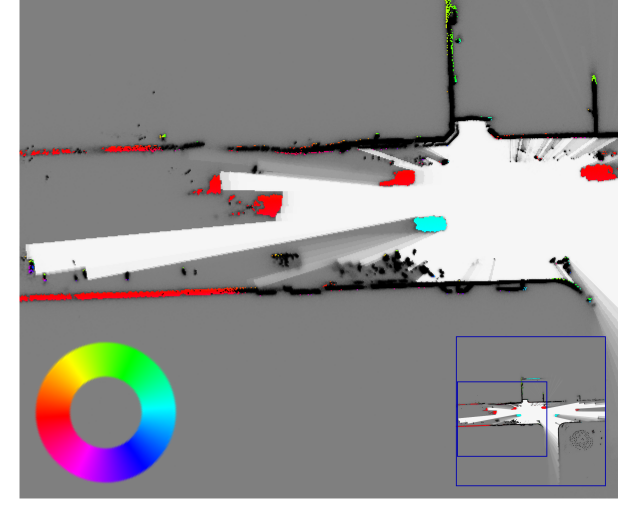}
		\subcaption{The orientation of the velocity estimates is visualized as colors, according the colored circle, for velocities $v\ts{m}>0.8$~m/s.}
		\label{fig:qualitativeEvalMoving}
	\end{subfigure}
	\caption{Qualitative comparison of our approach (left) with a particle-based grid map (right) with a stationary (a) and moving ego-vehicle (b). The images show a cutout of the grid map, according the illustration in the bottom right corner.}
	\label{fig:qualitativeEval}
	\vspace{-4mm}
\end{figure}
In order to evaluate the effectiveness of our ego-motion compensation, we divide the validation set in 4 sequences with a stationary ego-vehicle and 20 sequences, containing mainly a moving ego-vehicle. 
The results in Table \ref{tab:CellLevel} show, that the EPE is comparable in a stationary and a moving setting.
For both approaches, the mIoU which measures the performance of detecting static and dynamic cells is worse for the setting with a moving ego-vehicle.
However, the performance loss is only marginal in our approach.
The difficulty of detecting static cells in a setting with moving ego-vehicle is explainable with the changing field of view.
In Fig.~\ref{fig:qualitativeEvalMoving} a scenario is depicted, where a wall of a building sideways of the ego-vehicle, heading to the left, gets uncovered and is therefore difficult to detect as static area. 
Here, our approach is clearly superior compared to a particle-based approach as it can use context information to minimize incorrect velocity estimates in static environment.
\section{Conclusions}
In this work, we have presented a recurrent neural network architecture to solve the task of estimating DOGMs in various driving scenarios, using data from a moving ego-vehicle.
For this purpose, we introduce a method for ego-motion compensation in a fully convolutional network with recurrent layers on different scales.
Our evaluations show improved performance compared to the related work, especially in the separation of static and dynamic area in scenarios with a moving ego-vehicle.
In future work, we aim to include the prediction of instances with semantic classes in our DOGM.
%
%
%



%
%
%
%
%
%
\bibliographystyle{IEEEtran}
\bibliography{IEEEtranControl,IEEEabrv,mybibfile}
\end{document}